\documentclass[final]{cvpr}

\usepackage{times}
\usepackage{epsfig}
\usepackage{graphicx}
\usepackage{amsmath}
\usepackage{amssymb}
\usepackage[dvipsnames]{xcolor}

\usepackage{comment}
\usepackage{xparse}
\usepackage{xspace}
\usepackage{tabularx}
\usepackage{multirow}
\usepackage{caption}
\usepackage{float}
\usepackage{subcaption}
\usepackage{booktabs}

\usepackage{csquotes}

\usepackage[ruled,vlined]{algorithm2e}

\usepackage[pagebackref=true,breaklinks=true,colorlinks,bookmarks=false]{hyperref}

\begin{document}

\newcommand{\fulldataset}{VQA Perturbed Pairings\xspace}
\newcommand{\dataset}{VQA P2\xspace}
\newcommand{\ogvqa}{VQA v2\xspace}
\newcommand{\vqareph}{VQA-Rephrasings\xspace}
\newcommand{\framework}{Q3R\xspace}
\newcommand{\softmax}{\mathrm{softmax}}
\newcommand{\eqnref}[1]{Equation~\ref{#1}}
\newcommand{\appref}[1]{Appendix~\ref{#1}}
\newcommand{\norm}[1]{\left\lVert#1\right\rVert}
\newcommand{\rulesep}{\unskip\ \vrule\ }


\title{Learning from Lexical Perturbations for Consistent Visual Question Answering}

\author{
Spencer Whitehead$^{1}$, Hui Wu$^{2}$, Yi Ren Fung$^{1}$, Heng Ji$^{1}$, Rogerio Feris$^{2}$, Kate Saenko$^{2,3}$\\
$^{1}$UIUC \quad $^{2}$MIT-IBM Watson AI Lab, IBM Research \quad $^{3}$Boston University\\
{\tt\small \{srw5,yifung2,hengji\}@illinois.edu},
{\tt\small \{wuhu,rsferis\}@us.ibm.com},
{\tt\small saenko@bu.edu}
}

\maketitle

\begin{abstract}
Existing Visual Question Answering (VQA) models are often fragile and sensitive to input variations. 
In this paper, we propose a novel approach to address this issue based on modular networks, which creates two questions related by linguistic perturbations and regularizes the visual reasoning process between them to be consistent during training.
We show that our framework markedly improves consistency and generalization ability, demonstrating the value of controlled linguistic perturbations as a useful and currently underutilized training and regularization tool for VQA models. 
 We also present \fulldataset (\dataset), a new, low-cost benchmark and augmentation pipeline to create controllable linguistic variations of VQA questions. 
Our benchmark uniquely draws from large-scale linguistic resources, avoiding human
annotation effort while maintaining data quality compared to generative approaches.
We benchmark existing VQA models using \dataset and provide robustness analysis on each type of linguistic variation.\footnote{Data and resources are publicly available: \\ \url{https://github.com/SpencerWhitehead/vqap2}}
\end{abstract}

\section{Introduction}\label{sec:intro}
\begin{figure*}[t]
\small
    \centering
        \includegraphics[width=0.85\textwidth]{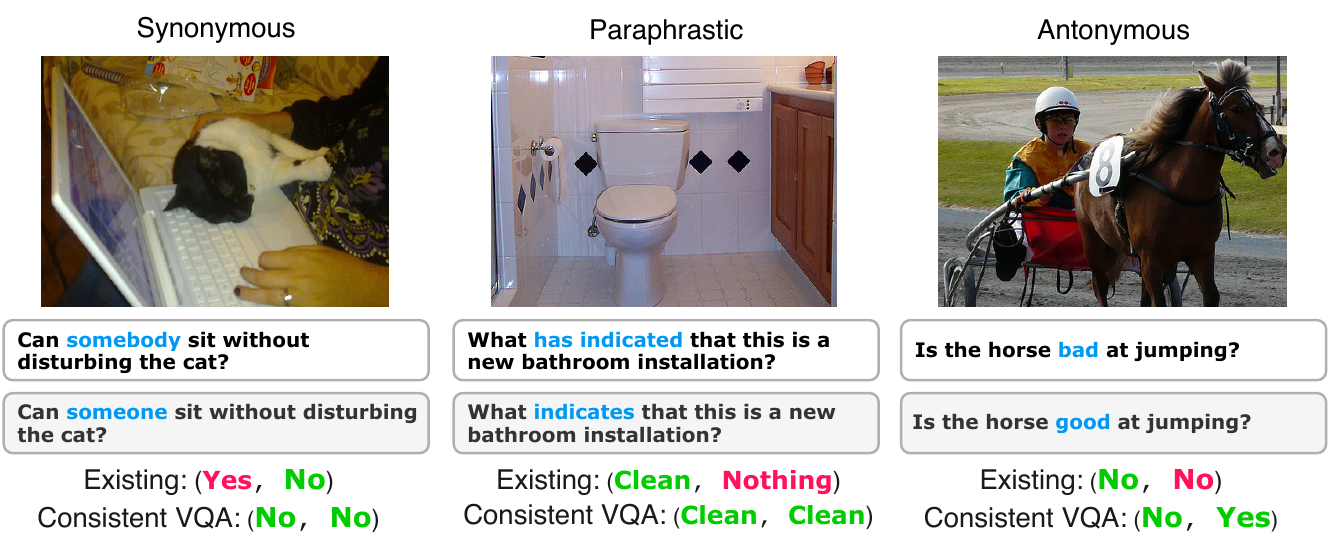}
\caption{
We show that existing VQA models can be surprisingly sensitive to simple, low-level linguistic changes. Examples show three types of lexical changes, along with answers predicted by a recent model~\cite{pythia18arxiv}. The goal of this 
work is twofold: how to benchmark such phenomenon, and
how to improve a model's robustness against linguistic perturbations. 
}
\label{fig:compare_examples}
\end{figure*}

Though great progress has been made towards Visual Question Answering (VQA), the approaches lack robustness and are sensitive to input variations~\cite{goyal2017vqav2,agrawal2018don,xu2018fooling,cycleconsist2019}. In
particular, prior work~\cite{cycleconsist2019} has shown that when presented with a  rephrased version of a question, VQA models often produce inconsistent answers. 
We conjecture that this is likely because most VQA approaches ignore interconnections between related questions and handle each example independently during training, even though learning such relationships is paramount to generalization, robustness, and compositionality. 

While a few attempts have been made to improve VQA robustness via augmentation and regularization techniques~\cite{wu2019self,cycleconsist2019,ray2019sunny}, they encourage consistency among related questions at the answer prediction level, without considering the stronger form of consistency between the intermediate computation steps. Furthermore, proposed augmentations are either costly (human-generated~\cite{cycleconsist2019}) or suffer from quality control issues~\cite{tang2020semanticeqadv}. 

In this work, we propose a novel robust VQA approach that first augments the question and then enforces consistency not only of the answer, but also of the intermediate representations computed by the model. The intuition is that (like humans) VQA models should follow the same reasoning steps to solve two differently-phrased questions that have the same meaning. For example, to answer both questions \textit{``Are the buildings tall?''} and \textit{``Are the buildings short?''} a model should follow the same process, \textit{i.e.} detect buildings in the image, then classify their height.

How to enforce such consistency of the intermediate reasoning steps appears to be a hard problem in general. We propose that an effective way to improve this stronger notion of \textit{reasoning consistency} is by maintaining the associated computation steps between questions that differ by controlled variations. We leverage the family of interpretable, compositional VQA models called \textit{Neural Module Networks (NMNs)}~\cite{andreas2016neural,hu2017n2nnmn,hu2018explainable,shi2019xnm} which explicitly represent sub-tasks like object detection and spatial reasoning as \textit{modules} within the network and predict sequences of weights over these modules (akin to \textit{programs}) to solve each question. However, unlike existing NMN work, we regularize the model to learn not only how to compose sub-tasks, but also follow the same sequences of sub-tasks to answer two variations of the same question, illustrated in \figref{fig:intuition}.

In addition, we also propose a novel data augmentation algorithm based on auxiliary linguistic knowledge freely available in text-only corpora. Examples like in~\figref{fig:compare_examples}, seem to suggest that existing VQA models ignore the relatedness between questions with mild linguistic modifications.
Questions like the antonymous pair in \figref{fig:compare_examples} are related in that they query the same property of the same object, but differ in that they ask the opposite of each other.
Existing work ignores such modifications and focuses either on human-written paraphrased VQA questions~\cite{cycleconsist2019} or auto-generated paraphrases using back-translation~\cite{tang2020semanticeqadv}.
Human paraphrases tend to add filler phrases or even change the question meaning (see \secref{sec:dataset}) and are very costly to annotate, while back-translations are also hard to control and have quality problems.
Motivated by this, we propose to augment data with changes at a low level, such as simple lexical substitutions. We create  variations  by substituting parts of the questions using rules extracted from large-scale language resources~\cite{miller1995wordnet,pavlick2015ppdb} which keeps the meaning and the realistic distribution of the original questions while avoiding the cost and/or semantic incoherence commonly found with prior work.

Finally, we contribute \fulldataset (\dataset), a dataset of perturbed questions derived from \ogvqa that can be used to 
measure robustness to the specific linguistic phenomena not previously evaluated in VQA literature, covering the usage of different synonyms (\emph{Synonymous}), different phrasings (\emph{Paraphrastic}), and opposite attributes (\emph{Antonymous}).
\dataset is comprised of 36.2k VQA questions, where each question has a corresponding question in \ogvqa~\cite{goyal2017vqav2} that it differs from by controlled perturbations.
\dataset can be easily expanded without the need for expensive human annotations~\cite{hudson2019gqa,ray2019sunny,cycleconsist2019}, while maintaining control over the  perturbations used.

To summarize, our main contributions are:
\begin{itemize}
  \item A novel VQA consistency regularization method that augments questions and enforces similar answers and reasoning steps for the original and augmented questions.
    \item A new data augmentation method and VQA robustness benchmark, \dataset, to diagnose the robustness of VQA models to controlled linguistic perturbations.
    To the best of our knowledge, this is the first approach to use large-scale linguistic resources automatically mined from human-written text to enhance VQA.
   \item Experiments applying our approach to regularize the training of modular, compositional, and interpretable VQA models, demonstrating that it leads to significant improvement in model consistency.
\end{itemize}

\section{Related Work}\label{sec:relwork}

\noindent\textbf{VQA.} 
Prodigious progress has been made towards VQA.
Models that learn strong feature backbones with co-attention/bilinear attention have achieved top scores on the widely adopted \ogvqa benchmark~\cite{anderson2018butd,kim2018bilinear}.
Recently, vision and language transformer models~\cite{tan2019lxmert,lu2019vilbert,su2019vl,li2019visualbert}, especially when pre-trained
on large-scale vision and language datasets, have led to state-of-the-art performance
on VQA accuracy. 
Despite this success, these models are generally not transparent and hard to diagnose.
In contrast, compositional and modular approaches, like NMNs, offer a number of distinct advantages, such as being more interpretable~\cite{hu2018explainable,mascharka2018}, requiring less training data~\cite{kim2018visual,mao2019neurosymbol}, and being able to handle complex, compositional questions~\cite{andreas2016neural,hu2017n2nnmn,hu2018explainable,shi2019xnm}.
Although, there is often a trade-off between accuracy and interpretability.

\noindent\textbf{VQA Benchmarks.} 
\ogvqa~\cite{goyal2017vqav2} has become the de facto training and evaluation data resource. Although improvements 
on \ogvqa test accuracy has marked encouraging progress on VQA, previous work has found that models are prone to learning 
superficial correlations by taking advantage of dataset 
bias~\cite{goyal2017vqav2,agrawal2018don}, and the standard VQA accuracy 
metric does not help to quantify such bias. A number of efforts have since 
been proposed to serve as additional diagnostic and evaluation tool for VQA.
CLEVR~\cite{johnson2017clevr} and GQA~\cite{hudson2019gqa} create compositional questions which are hard to answer if the model relies on language bias.
However, the questions in these datasets are synthetic and do not match the characteristics 
of natural questions about real images. VQA-CP~\cite{agrawal2018don} creates training and testing data splits which have different answer distributions. TDIUC~\cite{kafle2017analysis} categorizes and measures performance by 12 different question types. While these
benchmarks can measure a model's generalization ability to new question types or
new answer distributions, they do not serve to examine how much an answer might change if we change a specific input question. More related to our work is VQA-Rephrasings dataset~\cite{cycleconsist2019}, which has rephrased questions associated with a subset of original VQA questions.
However, as we have discussed, the open-endedness nature of human-written
paraphrases makes it hard to identify the source of model inconsistenties. 
Unlike our dataset (\dataset), it does not control for the 
specific type and degrees of question variations, making it hard to codify 
how questions are related to one another and learn from such relationships. 

\noindent\textbf{Improving VQA Robustness.} 
Robustness of deep learning models has been intensely studied in recent years,
both in the context of image classification~\cite{xie2020self,hendrycks2019augmix} and in a number of natural language understanding tasks~\cite{ganng2019,cheng2018towards}. 
Within VQA, robustness has been studied under the perspective of grounding model predictions to visual regions that are interpretable
and consistent with human annotated attention maps~\cite{li2018vqgvqa,wu2019self}.
While visual grounding provides a better basis for learning representations of concept
words in questions, many words and expressions in VQA questions do not have a direct 
correspondence with visual regions ({\sl i.e.,} not ``groundable'').
More related to our work, \cite{wu2019self,cycleconsist2019,ray2019sunny} examines model consistency under input question variations.
However, these models that consider consistency among related questions only do so at the answer prediction level~\cite{ray2019sunny,cycleconsist2019,wu2019self}.
Additionally, another approach applies embedding constraints between questions to improve performance~\cite{teney2019incorporating}.
Our framework not only employs answer prediction loss, but also dissects the model and enforces similarities on the sub-tasks involved, which aligns with compositional and transferable modeling of VQA.

\section{\dataset Benchmark}\label{sec:dataset}

Our goal is to create an objective benchmark to measure the progress of robustness in VQA models, specifically the consistency of VQA predictions under different linguistic perturbations of the input questions.
So the question is: what is the best method to create realistic, expressive, and semantically coherent, in an efficient fashion? 

Visual question generation methods~\cite{li2018vqgvqa,liu2018ivqa} can generate a variety of questions, but the output can be incoherent for a specific image or may be limited to certain types of questions~\cite{liu2018ivqa}. More recently, back-translation has emerged as a method of creating paraphrased questions for VQA~\cite{tang2020semanticeqadv}.
However, in general, controlled generation of text is a difficult open challenge~\cite{hu2017controlgen}.
Alternatively, template-based methods~\cite{hudson2019gqa,ray2019sunny} can offer more control and semantic coherence, but require extensive annotations ({\sl e.g.,} scene graphs~\cite{visualgenome}) and are limited in expressiveness by the templates used. Human-written paraphrases is 
the other option to generate robustness benchmark~\cite{cycleconsist2019}.
Other than the obvious drawback of being more costly, human-written 
question tend to be generic, not good for diagnosing the particular source
that causes answer prediction inconsistencies, as well as being often overly verbose (as shown in \secref{sec:othermethod}).

Different from existing approaches, we build our dataset from the original \ogvqa data, and create variations by substituting parts of the questions using rules extracted from large-scale language resources.
Doing so keeps the expressiveness and realistic distribution of the original questions and avoids the semantic incoherence commonly found with trained question generators. 
Utilizing large-scale linguistic resources offers an efficient means for creating questions that contain linguistic variations exhibited by humans~\cite{pavlick2015ppdb}.

\subsection{Paraphrasing Creation Pipeline}
\begin{table}[!ht]
\centering
\small
\begin{tabular}{l@{\hskip 0.03in}c} 
\toprule

\textbf{Perturbation} & \textbf{Example rules} (source $\to$ target) \\
& \\

\hline

\multirow{3}{*}{Synonymous} & \textit{car} $\to$ \textit{automobile} \\
& \textit{refrigerator} $\to$ \textit{fridge} \\
& \textit{phone} $\to$ \textit{telephone} \\

\hline

\multirow{3}{*}{Paraphrastic} & \textit{performed in} \texttt{[NP]} $\to$ \textit{carried out within} \texttt{[NP]} \\
& \textit{be considered} \texttt{[NP]} $\to$ \textit{be viewed as} \texttt{[NP]} \\
& \textit{participating in} \texttt{[VP]} $\to$ \textit{taking part in} \texttt{[VP]} \\

\hline

\multirow{3}{*}{Antonymous} & \textit{open} $\to$ \textit{closed} \\
& \textit{wet} $\to$ \textit{dry} \\
& \textit{full} $\to$ \textit{empty} \\

\bottomrule
\end{tabular}
\caption{Example rules used to generate the data. Synonymous and antonymous rules are single word, while paraphrastic rules can contain multi-word expressions and grammatical constraints.}
\label{tab:dataset_stats}
\end{table}

\noindent\textbf{Substitution Extraction.}
We extract lexical substitution rules from the Paraphrase Database 2.0 (PPDB)~\cite{pavlick2015ppdb}, a lexical database containing over 100 million paraphrases automatically mined from human-written text, as well as WordNet~\cite{miller1995wordnet}, two large-scale linguistic resources, and apply these rules to existing \ogvqa questions.
We create three types of perturbations:
1) \emph{synonymous} perturbations that substitute a single word with its synonym;
2) \emph{paraphrastic} perturbations that substitute multi-word phrases for single/multi-word phrases with the same meaning;
and 3) \emph{antonymous} perturbations where adjectives or verbs are substituted with their antonyms, which explores the ability to understand opposite states of an attribute or action.
We extract synonymous and paraphrastic rules from the lexical and syntactic subsets of PPDB respectively, and accept the rule if the target is ``equivalent'' or ``entailed'' by the source, according to PPDB's lexical constraints. 
We gather antonymous rules from WordNet.

\noindent\textbf{Rule Refinement.}
We automatically determine which rules can be applied by matching the source phrases and grammar requirements to the questions, discarding rules that are not applicable to any questions.
We ensure that our rules are not simply adding unknown words by removing rules whose source or target contain words that don't appear in the VQA vocabulary.
We filter the synonymous and paraphrastic rules with a minimum confidence threshold to prevent low quality/frequency substitutions.
Examples of these rules are shown in \tabref{tab:dataset_stats}, where each rule is a mapping from a source to target phrase.

\noindent\textbf{Applying Substitution Rules.}
We apply the filtered rules to \ogvqa questions and obtain our final benchmark.
Since consistency metric requires the groundtruth annotation for the each
question, and \ogvqa testing sets do not offer publicly available answer 
annotations, we use \ogvqa validation set as the basis for our benchmark. 
This practice is common among methods that require per-question answer annotations~\cite{cycleconsist2019,agrawal2018vqacp}. 

During the rule application process, for each question, multiple substitutions for a specific type of perturbation can be made, which increases variations.
For antonymous rules, we perform word sense disambiguation~\cite{pywsd14} and limit their application to yes/no questions of the form ``\textit{is/are the}'' and ``\textit{is/are this/these}'' where the WordNet synsets of the source word in both the question and rule match.
Antonymous rules are limited to these kinds of questions to help ensure that we only apply the antonym substitutions to attributes/states that are directly queried ({\sl e.g.,} ``\textit{Is the window open?}''), and not simply mentioned in the question ({\sl e.g.,} ``\textit{What's near the open window?}'').
For answers, synonymous and paraphrastic questions use the same answers as their \ogvqa counterparts as they share the same meaning and antonymous questions take the opposite answers. This results in a set of $36,175$ question-answer pairs that are perturbed counterparts of \ogvqa validation questions.

\begin{figure}[!ht]
\centering
\includegraphics[width=\columnwidth]{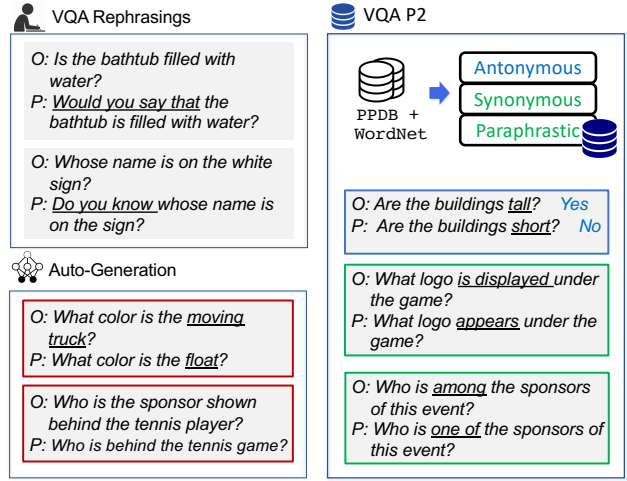} 
\caption{Examples of \dataset, human generated~\cite{cycleconsist2019}, and back-translated (Auto-Generation) variants (P) of input questions (O). \dataset uniquely offers three types of controlled linguistic variations to benchmark robustness.}
\label{fig:compare}
\end{figure}

\subsection{Comparison to Other Paraphrasing Methods}
\label{sec:othermethod}

Two main alternative methods for generating paraphrases of questions are human annotations and generative approaches.
\figref{fig:compare} shows a few examples from human-written paraphrased VQA questions~\cite{cycleconsist2019} and auto-generated paraphrases using back-translation~\cite{tang2020semanticeqadv}.
When primed with the original question, human annotators tend to give paraphrases that are longer than the original question, with a large percentage of these simply adding filler phrases that do not change the original sentence by much.
For example, in the first human-written example in \figref{fig:compare}, the original question is essentially still preserved within the paraphrased version.
Additionally, human paraphrases can introduce multiple sources of variations, such as introducing commonsense related items, sentence structural changes as well as lexical alterations.
Consequently, using human-written paraphrases as a diagnostic benchmark lacks a level of precision needed to diagnose and understand model performance.
On the other hand, automated methods like back-translation suffer from many quality control issues.
\figref{fig:compare} shows how this method frequently generates mismatched phrasal replacement or semantic drift, where the generated questions no longer hold the same meaning.

In contrast, our benchmark has built-in quality control as the replacement rules extracted from linguistic resources correlate well with human judgements~\cite{pavlick2015ppdb} and are generally reliable.
We target at linguistic perturbations to factor out other sources of variations when diagnosing models. 
Our dataset offers control over the kinds of perturbations used and allows for more precise evaluation of model consistency, meaning we can evaluate a model's capacity for addressing specific types of perturbations offering more diagnostic insight.
Additionally, due to the fast, automated nature of this process, our dataset is easily extensible to more linguistic variations and larger sets of perturbations.

\section{Approach}\label{sec:overview}

\begin{figure*}[!ht]
\centering
\begin{tabular}{ ccc }
\includegraphics[height=.27\textwidth]{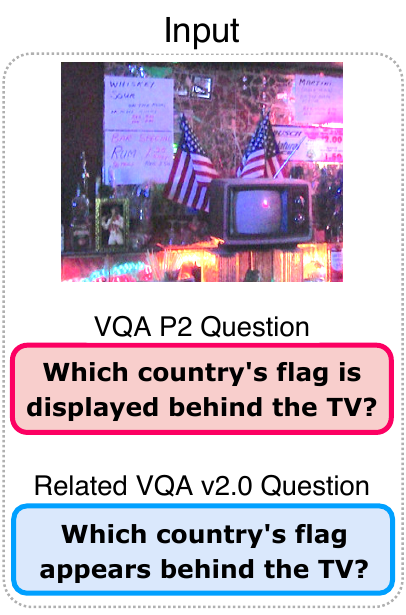} &
\quad
\includegraphics[width=.61\textwidth]{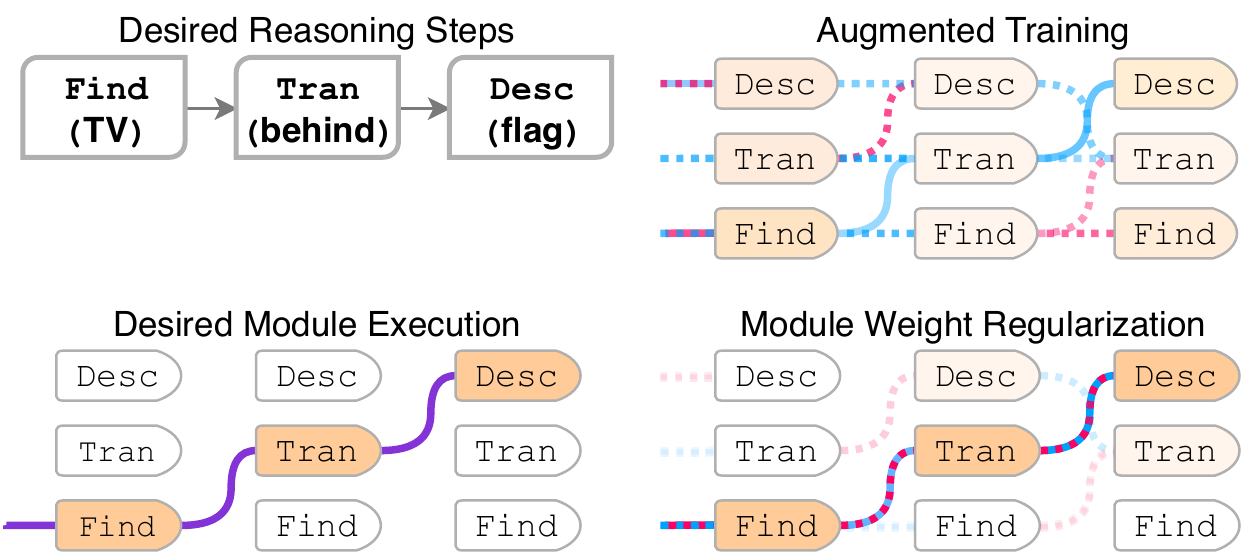} &
\end{tabular}
\caption{We propose a novel Q3R framework that improves the robustness of VQA models against linguistic variations by augmenting questions and encouraging similar module weights between related questions.}
\label{fig:intuition}
\end{figure*}

Given an image-question pair, $(I, Q)$,
a VQA model maps the pair to a distribution over an answer set,
$f(I, Q) \rightarrow \mathbf{a}$. Existing approaches most often treat this as a classification task, minimizing the prediction loss between the predicted answer and the ground truth answer.
This standard approach, however, does not take into account possible relations among questions.
The goal of our approach is to train a model to be aware of question relationships, and thereby learn to be more consistent when answering.

\noindent\textbf{Modeling Question Relatedness.}
Typically, given an input question, a predictable set of reasoning steps are expected to answer the question.
For example, in \figref{fig:intuition}, the sub-networks which can answer ``\textit{Which country's flag \underline{is displayed} behind the TV?}'' should be able to decompose this task into components such as ``\texttt{Find}(\textit{TV})'', ``\texttt{Transform}(\textit{Behind})'', and ``\texttt{Describe}(\textit{which country's flag})'' and learn to transfer all sub-networks to the question: ``\textit{Which country's flag \underline{appears} behind the TV?}''
Essentially, there should be a set of elementary operations where each one solves a less complex task than the original question and related questions can share these operations, following a similar order of execution.
Based on this intuition, we propose our \textbf{Question-Relatedness Regularized Reasoning} (\textbf{Q3R}) framework.

\subsection{Q3R Framework}\label{sec:nmn_arch}

Our framework is comprised of three components: 1) a method to create linguistic variations 
of input questions; 2) a compositional backbone model, guided by question-based 
module selection; and 3) a mechanism to enforce similarities of
related questions at the module level. 

\begin{figure}
\centering
\includegraphics[height=.20\textwidth]{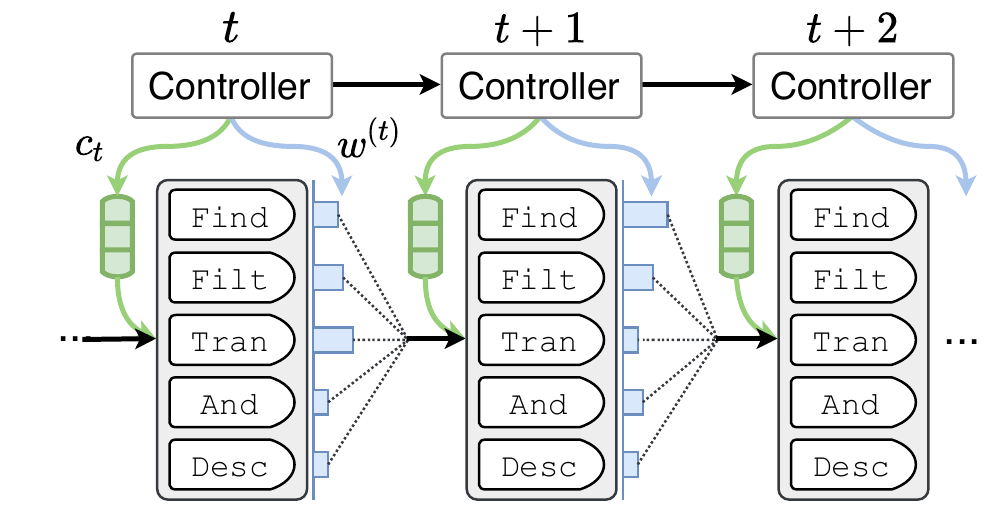}
\caption{Illustration of the backbone model.}
\label{fig:nmn_arch}
\end{figure}

\noindent\textbf{Creating Related Questions.}
The input to our pipeline is an image-question pair, $(I, Q)$, and, during training, a corresponding answer $A$.
We create a module, $g(Q,A) \rightarrow (\tilde{Q}, \tilde{A})$, that takes $Q$ and $A$ as input and outputs a related question-answer pair $(\tilde{Q}, \tilde{A})$.
In practice, any approach that can create linguistic variants of a given question-answer pair can be applied here.
For example, one can use generative models~\cite{tang2020semanticeqadv} to augment questions.
Although this data is typically too noisy for benchmarking purpose, it can provide additional data for training.
In this work, we simply use the rules provided by \dataset to create linguistic substitutions, making $g$ a function of the perturbation type, $t$, as well.
In future work, we will explore combinations of rule-based paraphrases and generative methods.

\noindent\textbf{Backbone Model.}
Our backbone model is comprised of input encoders, a controller network, and a set of re-usable modules, $\mathcal{M}$. The input encoders compute a set of visual and textual features for $I$ and $Q$, respectively.
As shown in \figref{fig:nmn_arch}, the controller is responsible for decomposing the reasoning process into a sequence of steps that are executed by the network.
At each step, $t$, the controller reads the question and, based on this question, produces module weights, $w^{(t)} \in \mathbb{R}^{|\mathcal{M}|}$, which are used for module selection, as well as a textual parameter, $c_t$, which 
is an input to the modules.
In this formulation, module selection is essentially an attention over the module outputs, which allows for end-to-end training.
The sequence of weights over all reasoning steps represents a soft layout that specifies what modules are utilized at each step.
The modules implement different sub-tasks that the model has at its disposal during reasoning.

Our framework is agnostic to the specific designs of the components of the backbone
network and can work within the general controller-module framework. We therefore select
two state-of-the-art compositional models to instantiate our backbone, specifically, 
\textbf{StackNMN}~\cite{hu2018explainable}, and \textbf{XNM}~\cite{shi2019xnm}. To 
verify that our training scheme is not tied to either one of the network, we also present
a new model by adopting modules from both networks, called \textbf{HybridNet}. 
More architectural details are in \appref{appendix:models}.

\noindent\textbf{Regularization Method.}
\label{sec:moduleloss}
We propose to regularize the training of the backbone compositional model to improve its consistency and robustness against linguistic variations at the module level. 
Controlling and regularizing question relatedness at the module level offers 
several benefits.
First, it provides finer control of the model's active sub-networks than only using supervision at the output layer.
Second, the related question pairs share intermediate activation similarity but do not necessarily need to match one another at a lower level ({\sl e.g.,} attention maps within modules), making the model less sensitive to surface-level sentence variation.

Given an image, $I_j$, as well as a pairing of an original question, $Q_j$, and its perturbed version, $\tilde{Q}_{j}$, the controller maps 
each question to a set of module weights at each reasoning step, $w_{j}^{(t)}$ and $\tilde{w}_{j}^{(t)}$, respectively.
Across all reasoning steps, these module weights can be interpreted as selecting ``paths'' over the grid of all modules across all time steps, especially when the weights are computed with Gumbel softmax~\cite{jang2016gumbelsoftmax} as done with XNM~\cite{shi2019xnm}.
Ideally, if two questions agree on the basic sub-tasks, they should also agree on the activated module paths (as in \figref{fig:intuition}).
Thus, we define the regularization loss term as:
\begin{align}\label{eqn:loss_func}
    \mathcal{L}_{m}(Q_j, \tilde{Q}_j, I_j) &= \lambda \sum_{t=1}^{T} d(w_j^{(t)}, \tilde{w}_j^{(t)}) ,
\end{align}
where $\lambda$ is a hyperparameter to scale the loss term and $d$ is a distance metric between two distributions.
We find that KL divergence or common vector norm losses, such as L1-norm, work well as $d$ in the proposed loss term.

\begin{algorithm}[h]
\small
\SetKwInOut{Input}{input}
\SetAlgoLined
\Input{steps $N$; input data $\mathcal{D}$; module $g$; network $f$}
 \For{$i\in\{1,...,N\}$}{
    sample $(I_i, Q_i, A_i)$ from $\mathcal{D}$ \\
    compute $\mathcal{L}_{CE}$ w.r.t. $(I_i, Q_i, A_i)$ \\
    \eIf{$\text{Bernoulli}(r) = 1$}{
        sample $(I_j, Q_j, A_j)$ from $\mathcal{D}$ \\
        $t \sim \text{cat}(\mathcal{T}|\mathbf{p})$ \\
        $(\tilde{Q}_j, \tilde{A_j}) \leftarrow g(Q_j, A_j, t)$ \\
        compute $\mathcal{L}_{m}$ w.r.t. $(I_j, Q_j, \tilde{Q}_j)$ \\
        $\mathcal{L} = \mathcal{L}_{CE} + \mathcal{L}_{m}$
    }{
        $\mathcal{L} = \mathcal{L}_{CE}$
    }
    update $f$ to minimize $\mathcal{L}$
 }
 \caption{Q3R Training Procedure}
 \label{alg:modulereg}
\end{algorithm}

\noindent\textbf{Training Procedure.}
Given the regularization term, $\mathcal{L}_{m}$, and the cross-entropy
loss between the predicted answer distribution and the ground truth answer prediction, $\mathcal{L}_{CE}(\mathbf{a}_i, \mathbf{a}_i^*|I_i, Q_i)$, we employ a multi-task training procedure~\cite{dong2015multi,luong2015multi}, where we treat each perturbation type as a different task, optimizing for each perturbation type individually.
As in Algorithm~\ref{alg:modulereg}, for each input
image-question pair sampled from \ogvqa, we compute $\mathcal{L}_{CE}$.
Then, with probability $r$, we sample a perturbed pairing to compute our module weight loss.
Otherwise, we simply update the model using $\mathcal{L}_{CE}$.
When utilizing our loss, we sample a perturbation type, $t$, from a categorical distribution, denoted $\text{cat}(\mathcal{T}|\mathbf{p})$, over the set of perturbation types, $\mathcal{T}$, with probabilities, $\mathbf{p}$.
We input the question $Q_j$ and the perturbation type into $g$ to obtain $\tilde{Q}_j$ that differs from the input question according to the sampled type.
We use the perturbed question and original image-question pair, $(I_j, Q_j, \tilde{Q}_j)$, to compute $\mathcal{L}_{m}$.
The network is then updated to minimize the sum of these loss terms.
In practice, when operating on batches, we sample a batch of a particular perturbation type.
We noticed that this procedure offers a more stable learning process and better performance than mixing different perturbation types in a single batch.

\section{Experiments}\label{sec:experiments}

\begin{table*}[ht]
\centering
\small
\begin{tabular}{lccccc@{\hskip 0.3in}ccc@{\hskip 0.3in}ccc}
\cmidrule[1pt]{1-12}
Model & Training Data & \ogvqa & \multicolumn{3}{@{\hskip 0.22in}l}{Synonymous} & \multicolumn{3}{@{\hskip 0.15in}l}{Paraphrastic} & \multicolumn{3}{@{\hskip 0.15in}l}{Antonymous}  \\
& & & \textit{Pert} & \textit{Ori} & $\Delta$ & \textit{Pert} & \textit{Ori} & $\Delta$ & \textit{Pert} & \textit{Ori} & $\Delta$ \\
\cmidrule[0.5pt]{1-12}
BAN~\cite{kim2018bilinear} & \ogvqa & 66.1 & 64.5 & 66.3 & -1.8 & 56.3 & 56.7 & -0.4 & 73.9 & 86.0 & -12.1 \\

\midrule[0.0001em]

Transformer~\cite{tan2019lxmert} & \ogvqa & 63.5 & 61.0 & 64.2 & -3.2 & 53.0 & 54.7 & -1.7 & 73.0 & 84.2 & -11.2 \\

\midrule[0.0001em]

StackNMN~\cite{hu2018explainable} & \ogvqa & 62.7 & 61.2 & 63.5 & -2.3 & 53.2 & 53.6 & -0.4 & 74.8 & 84.9 & -10.1 \\

\midrule[0.0001em]

XNM~\cite{shi2019xnm} & \ogvqa & 64.5 & 62.8 & 65.2 & -2.4 & 55.6 & 56.8 & -1.2 & 74.3 & 85.1 & -10.8 \\

\cmidrule[1pt]{1-12}
\end{tabular}
\caption{Accuracy on the \ogvqa validation set as well as \dataset subsets. \textit{Ori} and \textit{Pert} refer to the set of questions before and after applying a particular perturbation, respectively, and $\Delta$ is the difference in performance between these sets.}
\label{tab:accuracies}
\end{table*}

\noindent\textbf{Data and Metrics.}
We use \ogvqa (\appref{appendix:dataevals}) for training and \dataset for our main evaluations.
Since our goal is to benchmark and advance existing models' consistency, it requires groundtruth answer labels for each test question. 
Since answers for the \ogvqa test sets are not publicly available, we use questions from the validation set for testing and never use it during training, as is common for evaluating on annotated subsets of \ogvqa~\cite{agrawal2018vqacp,terao2020rephrasing}.
Our metrics are standard VQA accuracy as well as the consensus score (CS)~\cite{cycleconsist2019} between pairs of related questions.\footnote{A non-zero CS for a pair of questions requires a model to answer both questions correctly~\cite{cycleconsist2019}.}

\noindent\textbf{Models.} Existing VQA literature is vast~\cite{wu2017vqasurvey,teney2018tips,burns2019langfeatsmatter} and the goal of our work is not to exhaustively study VQA models.
We focus on the consistency and robustness of models in response to linguistically perturbed data.
Therefore, we pick representative models for our evaluations.
\textbf{BAN}~\cite{kim2018bilinear} is 
a bilinear model that achieves the single-model top performance on \ogvqa without external data.
BAN uses bilinear co-attention with residual connections to model the interactions between image regions and words.
\textbf{Transformer}~\cite{tan2019lxmert} is a top performing architecture that acts as a multi-modal encoder.
For fair comparison, we do not use large-scale pre-training.
As described in \secref{sec:nmn_arch}, \textbf{StackNMN}~\cite{hu2018explainable} and \textbf{XNM}~\cite{shi2019xnm} are examples of expert-free, end-to-end trainable NMNs.
We experiment with both models as well as a hybrid, called \textbf{HybridNet} (see \secref{sec:nmn_arch}).
Experiments with BAN and Transformer examine the consistency of non-NMN state-of-the-art architectures.
The NMN experiments explore the performance of different module implementations when employing our framework.

\noindent\textbf{Implementation.}
For fair comparison, all models use the same visual features from~\cite{anderson2018butd}, input textual features from~\cite{pennington2014glove}, and random seed of 0.
Whenever possible, we use publicly available implementations.
For \framework, $\lambda = 1.0$ and $r=0.2$ for all experiments and we use a L1 loss for our distance metric.
Please see \appref{appendix:implementation} for details.

\subsection{Benchmarking Robustness with VQA-P2}
We benchmark existing models on their robustness against controlled lexical 
perturbations, as shown in \tabref{tab:accuracies}.
Interestingly, we see that the different classes of models have less trouble with paraphrastic changes than they do with synonymous, with the average drop in performance being $-0.9$ compared to $-2.4$.
This is likely due to the fact that paraphrastic changes tend to effect transitional phrases ({\sl e.g.,} ``\textit{be considered}''), which models may ignore~\cite{subramanian2019analyzing}, whereas synonymous changes effect these as well as concept mentions ({\sl e.g.,} ``\textit{car}'') that are needed to answer the question.
We see that models struggle the most with antonymous changes, dropping at least $-10.1$.
Despite having lower \ogvqa accuracy, the NMN architectures perform better on the perturbed antonymous questions compared to BAN and Transformer.
The results suggest that the Transformer architecture~\cite{tan2019lxmert} trained from scratch is one of the less robust architectures across the different types of perturbations.
Overall, we see that existing models struggle with these controlled variations across the board, where the largest difficulties appear on the logical consistency measured with antonymous perturbations and concept mention consistency measured with synonymous perturbations.
To our knowledge, this is the first study of
VQA robustness analysis on different types of linguistic variations,

\begin{figure*}[t]
	\centering
	\includegraphics[width=0.85\textwidth]{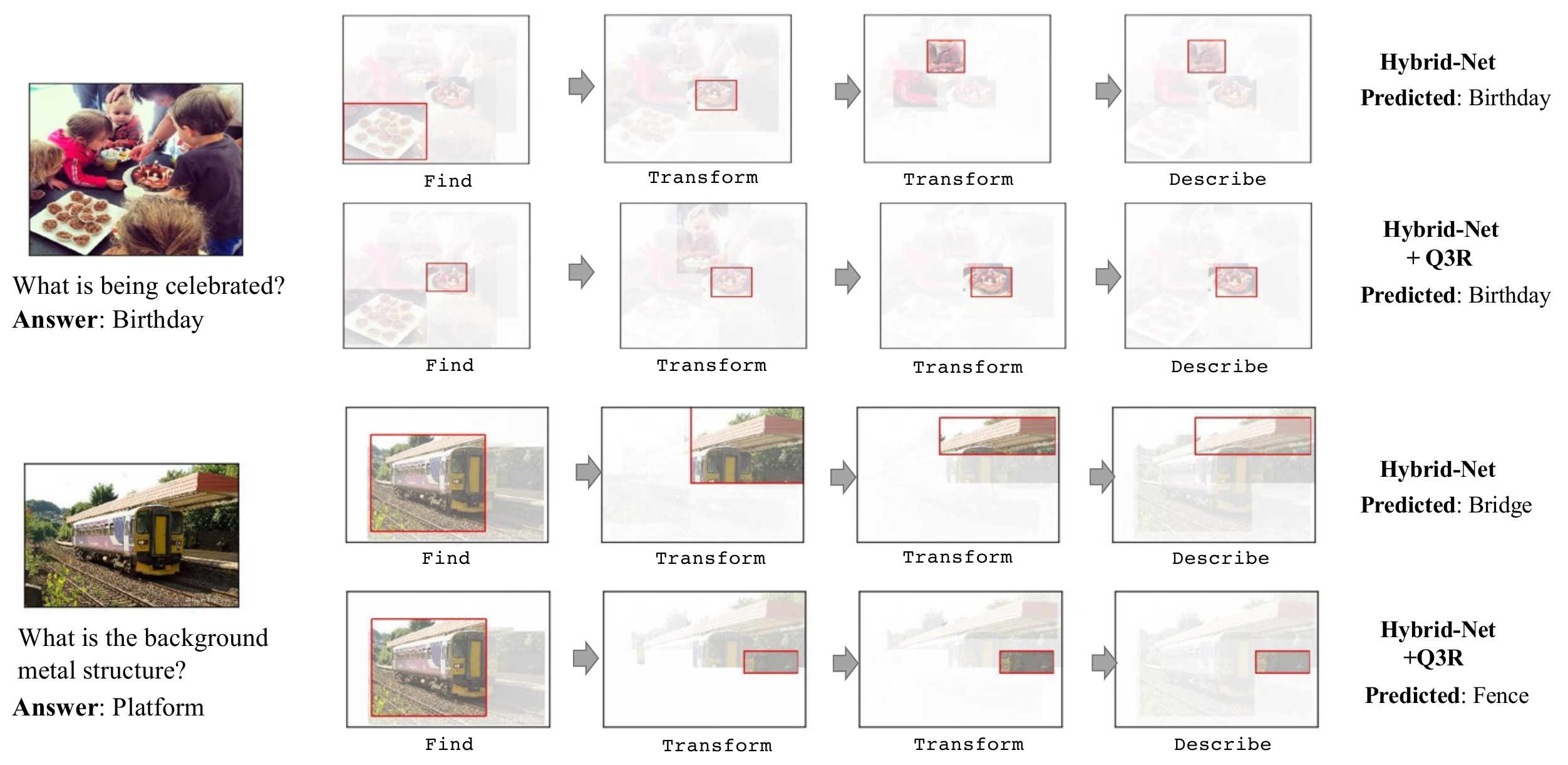}
	\caption{Visual attentions and selected modules across reasoning steps.
	Attention weights are indicated by the transparency of the bounding box; red boxes are the highest weighted regions. Please see \appref{appendix:examples} for more examples.}
	\label{fig:attention}
\end{figure*}

\subsection{Improving Model Robustness using \framework}

\begin{table}[ht]
\small
\centering
\begin{tabular}{lccc|c}
\cmidrule[1pt]{1-5}
   & \multicolumn{3}{c|}{\dataset} & \ogvqa \\
Model   & \textit{Pert} & \textit{Ori} & CS & \\

\cmidrule[0.5pt]{1-5}
StackNMN  & 63.3 &  66.9 & 66.2 & 62.6\\
\hspace{0.05in} +\framework & \textbf{66.9} & \textbf{67.4} & \textbf{72.2} & \textbf{62.7} \\

\midrule[0.0001em]

HybridNet & 63.3  & 67.0  & 66.6  & 63.0 \\
\hspace{0.05in}+\framework   & \textbf{67.0} & \textbf{67.4} & \textbf{72.5} & \textbf{63.1} \\

\midrule[0.0001em]

XNM  & 64.7 & 68.3 & 68.8 & 64.5 \\
\hspace{0.05in}+\framework & \textbf{68.1} & \textbf{68.9} & \textbf{74.4} & \textbf{64.7} \\

\cmidrule[1pt]{1-5}
\end{tabular}
\caption{Effect of adding \framework to different NMNs, measured by accuracy and CS. \textit{Pert} denotes perturbed questions; \textit{Ori} are the corresponding questions in \ogvqa.}
\label{tab:regularization_p2_acc}
\end{table}

We evaluate the effect of adding \framework to different NMN architectures. \tabref{tab:regularization_p2_acc} shows that, 
on \dataset, adding our framework (+\framework) results in significant improvements over the base models for both accuracy and CS.
We also measure validation accuracy on \ogvqa in \tabref{tab:regularization_p2_acc}, which shows modest yet consistent gains, although our focus is on consistency, not overall \ogvqa accuracy.
This shows that models regularized by \framework generally see improvements in consistency and overall performance, including on \ogvqa.

\noindent\textbf{Analysis by Perturbation Types.}
A benefit of having information on the specific types of linguistic variations is that we can profile a model's performance by each type to assist understanding and diagnosis.
\tabref{tab:perturbation_types} shows that models are generally more confident with the antonymous perturbations, likely because ``yes/no'' questions have higher answer prediction scores in general.
We note that performance gains are significant on single-word changes and less obvious on multi-word changes.

\begin{table}[ht]
\small
\centering
\begin{tabular}{p{1.5cm}cc|cc|cc}

\cmidrule[1pt]{1-7}
Model & \multicolumn{2}{c|}{Syn.} & \multicolumn{2}{c|}{Par.} & \multicolumn{2}{c}{Ant.} \\
  &  $\mu$ & CS & $\mu$ & CS & $\mu$ & CS \\
\cmidrule[0.5pt]{1-7}

StackNMN & 62.6  & 64.7 & 51.6 & 53.8 & 79.8 & 76.1  \\
\hspace{0.05in}+\framework & \textbf{64.0} & \textbf{70.3} & \textbf{52.9} & \textbf{56.9}  & \textbf{84.9}  & \textbf{84.7} \\

\midrule[0.0001em]

HybridNet & 62.5 & 65.0 & 53.3 & 55.7 & 79.9  & 76.4 \\
\hspace{0.05in}+\framework  & \textbf{64.1} & \textbf{70.8} & \textbf{55.0} & \textbf{59.4} & \textbf{84.0}  & \textbf{83.7} \\

\midrule[0.0001em]

XNM  & 64.1 & 67.6 & \textbf{56.7} & 60.7 & 79.7 & 76.0 \\
\hspace{0.05in}+\framework & \textbf{65.7} & \textbf{72.9} & 56.4 & \textbf{61.8} & \textbf{84.4} & \textbf{84.7} \\

\cmidrule[1pt]{1-7}
\end{tabular}
\caption{Average accuracy ($\mu$) between \textit{Pert}+\textit{Ori} and CS on each type of perturbation: Synonymous (Syn.), Paraphrastic (Par.) and Antonymous (Ant.).}
\label{tab:perturbation_types}
\end{table}

\subsection{Further Analysis}
\noindent\textbf{Effect of Regularization.}
To further understand the effect of applying our loss on the reasoning steps involved in executing NMNs, we compute different statistics using HybridNet.
For each question in \dataset, we compute the average normed difference in output module weights between the perturbed question, $\tilde{Q}_j$, and the original question, $Q_j$ for all steps.
Applying our loss leads to a $93\%$ reduction ($7.4\times10^{-3}$ to $0.5\times10^{-3}$)  on the average difference of the module weight distribution at each step. In other words, our framework encourages more consistent reasoning steps
between related questions. Also, it appears that our framework can lead to more intuitive visual attentions, as shown in \figref{fig:attention}. These results 
suggest that our framework may give rise to more interpretable and consistent
visual reasoning.

\noindent\textbf{Additional Evaluation.}
\label{subsec:rephrasings}
As we discussed in \secref{sec:othermethod}, human-written rephrasings typically contain various sources of change: such as those that involve common sense knowledge, or structural level change of sentences.
Nonetheless, we are interested in 
observing the outcome of the additional study on \vqareph. 
When trained with our Q3R framework, XNM receives a $+0.7$ accuracy improvement 
on paraphrased questions as well as a $+0.4$ improvement on CS score. 
This moderate gain may be explained by the fact that linguistic perturbations 
are present in human-written questions, so the model's robustness on this dataset
benefits from our training framework.

\subsection{Discussion}
\noindent\textbf{Comparison to Expert Layout Supervision.}
Many NMNs adopt expert layouts to guide the search for optimal module paths~\cite{andreas2016neural,hu2017n2nnmn}.
While this approach is suitable on synthetic datasets with simple scenes and spatial reasoning~\cite{johnson2017clevr}, it has limited success on realistic images~\cite{hu2018explainable}.
Our loss can be viewed as providing weak supervision to module layouts, avoiding the need for ground-truth module layout annotation, which is costly and not
clearly defined for natural questions about real images. 

\noindent\textbf{Beyond NMNs.}
Our results seem to suggest the advantage of representing and incorporating 
inductive bias at the modular level, rather than just using 
answer-level supervision.
While we have demonstrated our method using Neural Module Networks,  we note that it could be generalized to improve the robustness of any other interpretable VQA architecture that involves the computation of sub-tasks, such as those based on executable symbolic programs~\cite{yi2018neural,mao2019neurosymbol}.

\section{Conclusion}
We show that a promising direction to improve the robustness and consistency of VQA models is by modeling and learning from lexical perturbations.
We propose a novel approach based on modular networks, which creates two questions related by linguistic perturbation and regularizes the visual reasoning process between them to be consistent during training.
We introduce a new benchmark, \dataset, that features categorized, controllable linguistic variations that allows us to investigate and diagnose sources of inconsistencies in model predictions, will be made publicly available.
Empirical results show that existing models have difficulties with different types of linguistic variations and that our approach is effective towards improving robustness and generalization ability.

\clearpage
{\small
\bibliographystyle{ieee_fullname}
\bibliography{refs}
}

\newpage
\clearpage
\begin{center}
{\bf {\Large Appendix\\} }
\end{center}
\appendix

\setcounter{page}{1}
\section{Implementation Details}\label{appendix:implementation}

As noted in \secref{sec:experiments}, all models use the same visual features from \cite{anderson2018butd}\footnote{\url{https://github.com/peteanderson80/bottom-up-attention}} and same pre-trained GloVe embeddings from~\cite{pennington2014glove}\footnote{Common Crawl 840B: \url{https://nlp.stanford.edu/projects/glove/}}.
XNM, StackNMN, and HybridNet use the implementation provided by \cite{shi2019xnm} to ensure consistency amongst different NMN models.\footnote{\url{https://github.com/shijx12/XNM-Net}}
We implement the modules and controller of StackNMN to match the paper description and official implementation.
All NMN models are trained with the Adam optimizer~\cite{kingma2014adam} and have the same learning rate of 0.0008, training initialization seed of 0, and batch size of 256.
Following their implementations, we use hidden dimension sizes of 512 for StackNMN and 1024 for XNM, while we use 1024 for HybridNet to match XNM.
We use the recommended number of reasoning steps, $T=3$, for XNM and use the same for StackNMN.
HybridNet uses $T=4$ to test longer reasoning sequences as well as for visualization purposes.
With our \framework framework, $\lambda=1.0$ and $r=0.2$ for all experiments.
For BAN, we use the original source code
with the 8-glimpse model, provided by the authors and adopt their training settings.\footnote{\url{https://github.com/jnhwkim/ban-vqa}}
We do not use the counting module~\cite{zhang2018vqacount} nor additional training data from Visual Genome~\cite{visualgenome}.
For the Transformer model, we use LXMERT~\cite{tan2019lxmert} as the architecture and utilize the publicly available code with the recommended settings from the authors.\footnote{\url{https://github.com/airsplay/lxmert}}
For fair comparison, we do not use large-scale pre-training.

\section{Effect of Hyper-parameters}\label{appendix:hyperparams}

The main hyperparameters involved with using \framework are the weight parameter, $\lambda$, and the distance function, $D(\cdot)$.
Changing $\lambda$ from 1 to 0.2 results in
a value change of $<0.1$ on \ogvqa accuracy and $<0.1$ on CS score on \dataset, for StackNMN. Also, 
the difference of applying KL-divergence based loss function and L1-norm based loss function 
results is also $<0.1$, for both \ogvqa accuracy and CS score on \dataset.

\begin{table*}[ht]
\small
\centering
\begin{tabular}{ |c|c|c|c|c| } 
\hline
Module & Input & Output & StackNMN~\cite{hu2018explainable} & XNM~\cite{shi2019xnm} \\
\hline

\texttt{Find} & $x, c_t$ & $\hat{a}$ & $\text{conv}_2(\text{conv}_1(x) \odot Wc_t)$ & $ \text{conv}(g(x, W c_t))$ \\ 

\texttt{Transform} & $x, c_t, a$ & $\hat{a}$ & $\text{conv}_2(\text{conv}_1(x) \odot W_1 \sum(a \odot x) \odot W_2 c_t)$ & $\text{norm}((\sigma(R W c_t) \odot M) a)$ \\

\texttt{Filter} & $x, c_t, a$ & $\hat{a}$ & $\texttt{And}(\texttt{Find}(x, c_t), a)$ & (same as StackNMN) \\ 

\texttt{And} & $a_1$, $a_2$ & $\hat{a}$ & $\min(a_1,a_2)$ & (same as StackNMN) \\ 

\texttt{Describe} & $x, c_t, a$ & $z^{(t)}$ & $W_{1}^{\top}(W_2 \sum(a \odot x) \odot W_3 c_t)$  & $\sum a \odot x$ \\ 

\texttt{NoOp} & none & - & - & - \\ 

\hline
\end{tabular}
\caption{Neural modules of StackNMN and XNM. Here $x$ are the visual features, $c_t$ is the textual parameter, each $a$ (and $\hat{a}$) is an attention map over the image regions, $z^{(t)}$ is used to compute the final answer prediction, and all $W$ are learned parameters. For XNM, $R$ is the set of edge features and $M$ is the adjacency matrix of the scene graph.}
\label{tab:modules}
\end{table*}

\section{Additional Dataset Information}\label{appendix:dataevals}
To measure the model's robustness against question variations, it requires
the availability of answer annotations during evaluation. Since the testing data
of \ogvqa~\cite{goyal2017vqav2} does not have public groundtruth information, we use the validation split
of \ogvqa as the testing set for all models. For training, we use the 
training data of \ogvqa, which contains $443,757$ questions for $82,783$ images; for testing, we use the validation split of \ogvqa ($214,354$ questions for $40,504$ images) and \dataset ($36,175$ question-answer pairs)

In \secref{subsec:rephrasings}, we evaluate our framework on \vqareph~\cite{cycleconsist2019}. \vqareph is a human-written paraphrased dataset that spans the $40,504$ validation images of \ogvqa, where each image has a corresponding question group that contains $1$ original question from the \ogvqa validation set and $3$ rephrasings of that questions, for a total of $162,016$ questions.

\section{Model Details}\label{appendix:models}

\subsection{Controller-Module Architecture}\label{appendix:module_design}
Our proposed training framework is applicable to the any controller-module framework~\cite{andreas2016neural,hu2017n2nnmn,hu2018explainable,shi2019xnm}.
Next, we provide details of the components of our backbone architecture, which is comprised of input encoders, a controller, 
and a set of functional modules. 

\noindent\textbf{Input encoders.} 
The visual features, $x \in \mathbb{R}^{d_v\times K}$, represent $K$ different regions of the image, which may be grid features from a convolutional layer of a CNN~\cite{simonyan2014very,he2016resnet}, object features from an object detector~\cite{ren2015fasterrcnn}, or a scene graph~\cite{visualgenome,johnson2017clevr}.
Similarly, representations of question tokens, $(h_1,...,h_N)$, or the entire question, $q$, may be obtained via embeddings~\cite{mikolov2013efficient,pennington2014glove}, an encoder~\cite{hochreiter1997long,cho-etal-2014-learning}, or a pre-trained language model~\cite{devlin2019bert}.

\begin{figure}[ht]
\centering
\includegraphics[height=.25\textwidth]{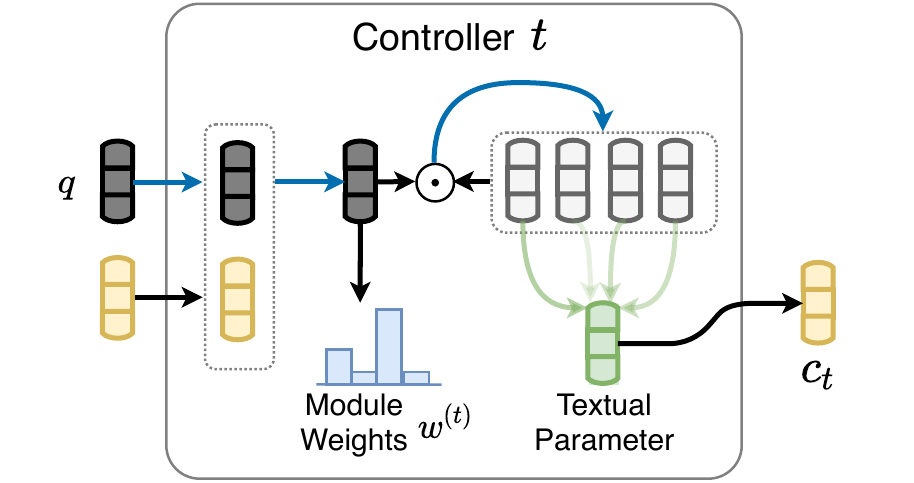}
\caption{Illustration of the controller network.}
\label{fig:controller}
\end{figure}

\noindent\textbf{Controller.} At each reasoning step, $t$, the controller reads the question and computes $c_t$ and $w^{(t)}$.
$w^{(t)} = \{w_{m}^{(t)}\}_{m=1}^{|\mathcal{M}|}$ is a set of module weights that are utilized by the network to perform a soft selection of modules at step $t$.
These module weights are computed via a soft attention~\cite{bahdanau2015neural}, forming a distribution over the modules, which is used to compute the output at step $t$ by taking the weighted average over each module's output.
The vector $c_t \in \mathbb{R}^{d}$ is a textual parameter, typically computed using an attention over the question tokens~\cite{hu2018explainable}, which is given as input to the modules and used, for example, as a query for visual attentions.
\figref{fig:controller} shows the architecture of the controller, which has
a similar design as the controller in~\cite{hu2018explainable,hudson2018compositional}.
At each step, the controller first concatenates the history vector, which is the textual parameter from the previous step, with the question representation, $q$, which is the final hidden state of a bi-directional GRU~\cite{cho-etal-2014-learning}.
Next, the concatenated vector is given as input to a MLP.
The output of the MLP (with an optional normalization layer) is then used to predict module weights as well as compute an attention over the question tokens, which yields a new textual parameter, $c_t$.

\noindent\textbf{Modules.}
Each module is responsible for an elementary operation, with inputs that may be
any combination of attention maps generated at previous steps, the textual parameter from the controller, the visual features, or any other relevant features.
A module outputs either an attention map over the image regions, $a \in \mathbb{R}^{K}$, or a vector, $z^{(t)} \in \mathbb{R}^{d}$, which may be used for answer prediction.
The output of the modules at the last time step is combined with the question representation and used to predict the final answer via an output layer, such as a MLP.

\subsection{Models}\label{appendix:base_models}
The backbone architectures can be instantiated differently to
realize a variety of models with distinct module functionalities, reasoning steps, feature backbones, etc.
In this work, we adopt three designs of the controller-module models to test the effects of our \framework training framework.

\noindent\textbf{StackNMN}~\cite{hu2018explainable}\textbf{.}
For StackNMN, the original method uses grid features from a CNN~\cite{he2016resnet} as visual features. For fair comparison with other
models, we use object features~\cite{anderson2018butd} for StackNMN, which is a stronger feature backbone for VQA.
The modules in this architecture largely use elementwise multiplications to fuse visual and linguistic features, compute different attention maps, or obtain answer vectors.
Additionally, in \texttt{Find} and \texttt{Transform}, 1D convolutions are also used to compute weighted visual and multi-modal features.
The specific module designs for StackNMN are shown in \tabref{tab:modules}.

\noindent\textbf{XNM}~\cite{shi2019xnm}\textbf{.}
For XNM, the visual features, $x \in \mathbb{R}^{d_v\times K}$, are object features that represent nodes in the input scene graph.
The edge features, $R \in \mathbb{R}^{K \times K \times 2d_v}$, for the scene graph are the concatenations of neighboring edges and $M \in \mathbb{R}^{K\times K}$ is the adjacency matrix of the scene graph.
The \texttt{Transform} module is the only module that considers the scene graph and uses the graph information to shift the visual attention according to the graph connectivity.
These modules adopt the naming conventions of StackNMN, but XNM, in particular, has a different \texttt{Transform} implementation, which learns attention transforms on image scene graph representations, and an alternative multi-modal fusion method~\cite{zhang2018vqacount}. The detailed module information is 
shown in \tabref{tab:modules}.

\noindent\textbf{HybridNet.}
To further investigate whether our framework can be effective regardless of the module implementations, we present a hybrid of StackNMN and XNM.
Specifically, HybridNet utilizes the \texttt{Transform} module of StackNMN, while maintaining the rest of the design from XNM.

\section{Output Examples}\label{appendix:examples}
\begin{figure*}[ht]
\centering
\includegraphics[height=.45\textwidth]{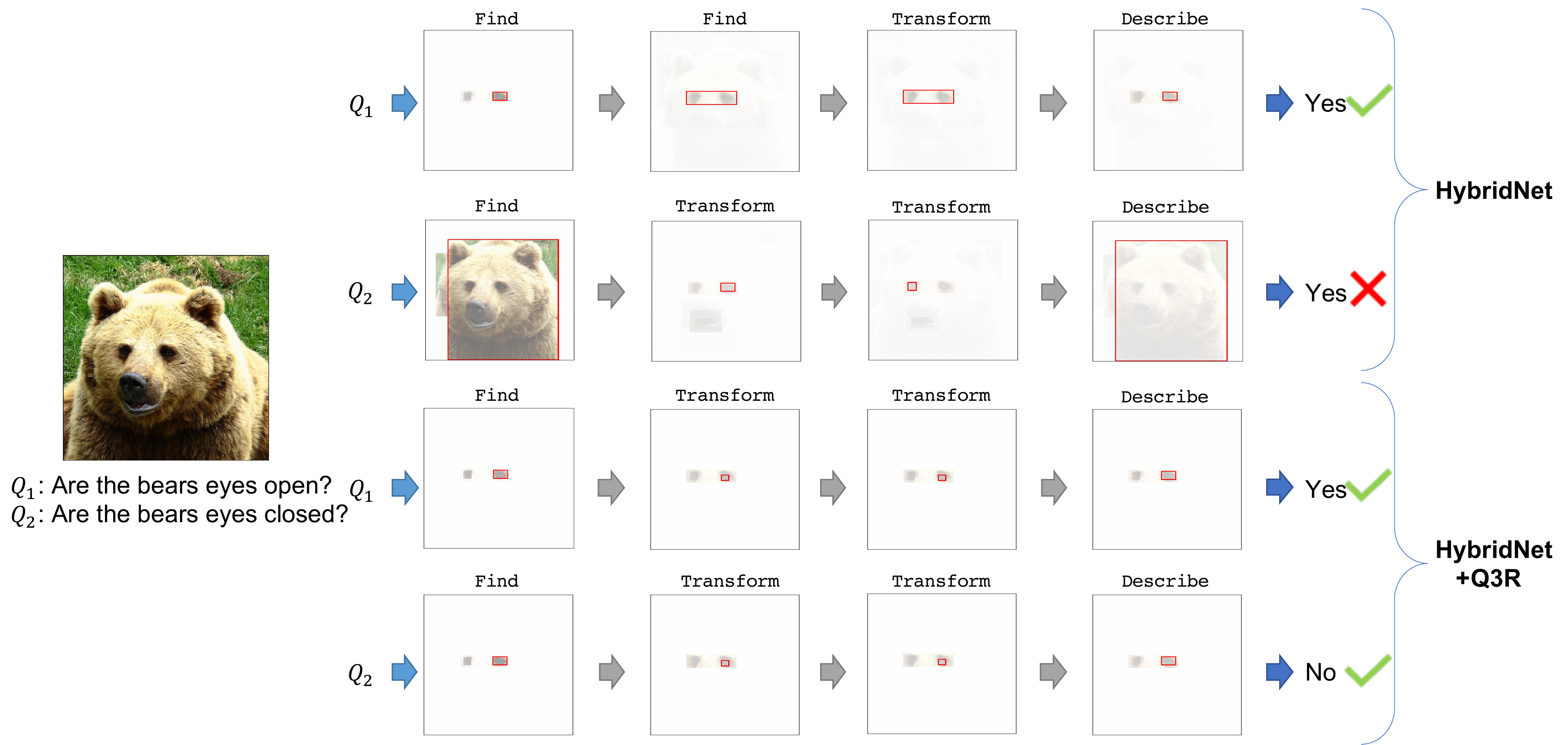}
\caption{Example visual attentions, module selections, and outputs for an antonymous pair of questions. The top two rows are from HybridNet trained only on \ogvqa, while the bottom two rows are from HybridNet trained with \framework.}
\label{fig:output_bear}
\end{figure*}

\begin{figure*}[ht]
\centering
\includegraphics[height=.45\textwidth]{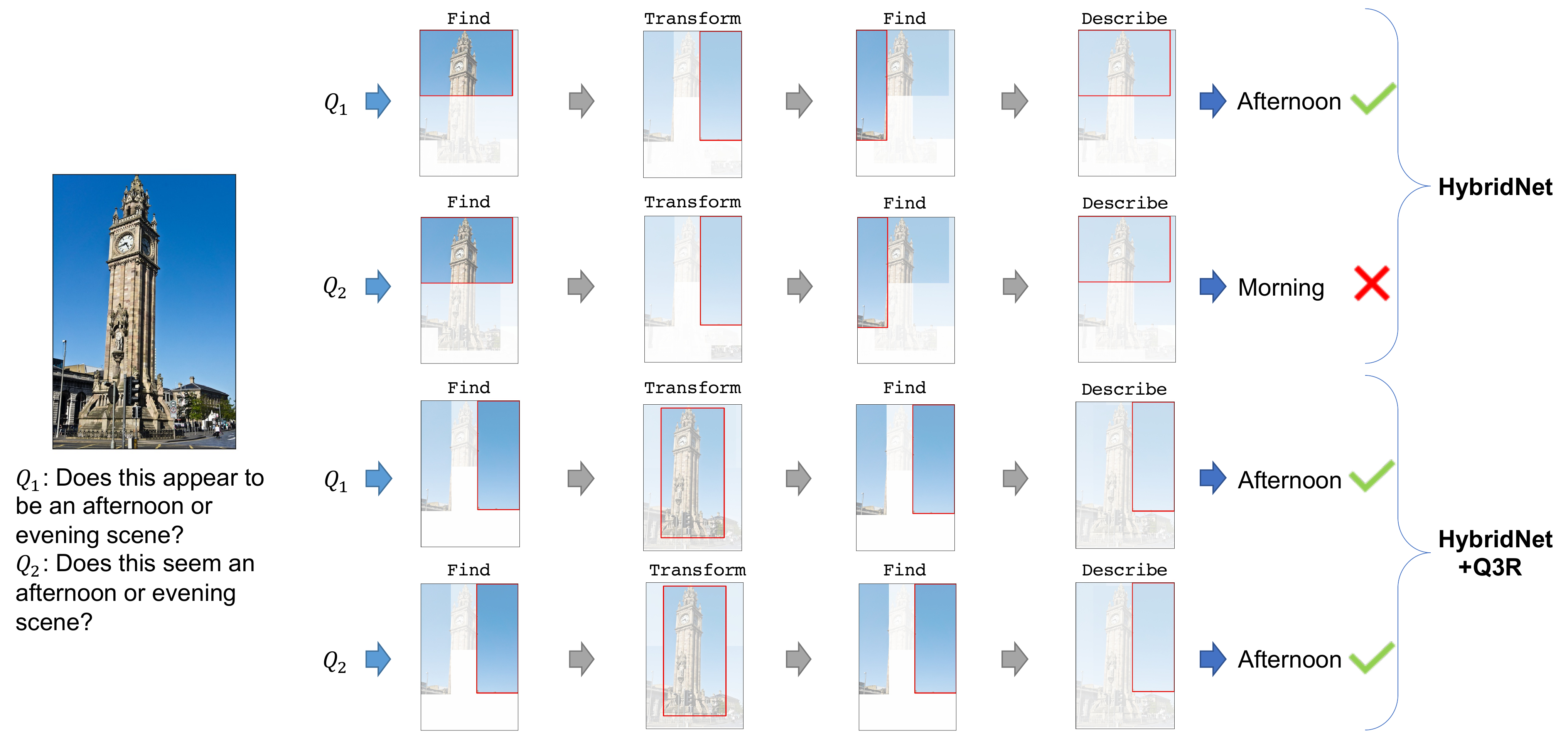}
\caption{Example reasoning outputs for a paraphrastic pair of questions. Again, the top two rows are from HybridNet trained on \ogvqa and the bottom two are from HybridNet+\framework.}
\label{fig:output_scenetime}
\end{figure*}

Here are some example outputs from HybridNet with and without \framework.
In \figref{fig:output_bear}, the model trained without \framework maintains its answers despite the perturbation, whereas the model trained with our framework predicts the appropriate answer and also matches visual attentions between them.
Then, in \figref{fig:output_scenetime}, we see an example where the questions share the same meaning and the same modules are selected for both questions and both models, but the model without our framework yields inconsistent answers.

\end{document}